%% file: root.tex
\documentclass[letterpaper, 10 pt, conference]{ieeeconf}  
\IEEEoverridecommandlockouts                            
\usepackage[utf8]{inputenc}
\usepackage[T1]{fontenc}
\usepackage{amsmath}
\usepackage{upgreek} 
\usepackage{bbm}
\usepackage{amsfonts}
\usepackage{bm}
\usepackage[noadjust]{cite}
\usepackage{graphics}
\usepackage{amssymb}
\usepackage{mathrsfs}
\usepackage[mathscr]{euscript}
\usepackage{threeparttable}

\makeatletter
\let\NAT@parse\undefined
\makeatother
\usepackage{graphicx}
\usepackage{soul, color, xcolor}
\usepackage[colorlinks, linkcolor=blue, anchorcolor=blue, citecolor=blue]{hyperref}

\definecolor{green}{HTML}{369b36}
\definecolor{yellow}{HTML}{ffc000}
\definecolor{LATblue}{HTML}{4472c4}  
\definecolor{cyan}{HTML}{00ffff}  
\definecolor{orange}{HTML}{ff9900}  
\definecolor{light_aquamarine}{HTML}{80E5BF }  
\definecolor{light_green}{HTML}{90ee90 }  
\definecolor{Jasmine}{HTML}{ffd580}

\title{\LARGE \bf
Robust and Accurate Multi-view 2D/3D Image Registration with Differentiable X-ray Rendering and Dual Cross-view Constraints
}

\author{Yuxin Cui, Rui Song, Yibin Li, Max Q.-H. Meng, \textit{IEEE Fellow}, Zhe Min$^{\ast}$
\thanks{$^\ast$ Corresponding Author.}% <-this % stops a space
\thanks{This work was supported by the National Natural Science Foundation of China under Grant 62303275, the National Natural Science Fund for Excellent Young Scientists Fund Program (Overseas) under Grant 22IAA01849, Jinan Municipal Bureau of Science and Technology under Grant 202333011.}
\thanks{Yuxin Cui is with the Department of Information Science and Engineering, Shandong University, Qingdao, China.}%
\thanks{Zhe Min, Rui Song and Yinbin Li are with the School of Control Science and Engineering, Shangdong University, Jinan, China. Zhe Min is also with UCL Hawkes Institute and the Department of Medical Physics \& Biomedical Engineering, University College London, London, UK.
{\tt\small minzhe@sdu.edu.cn}}%
\thanks{Max Q.-H. Meng is with the Shenzhen Key Laboratory of Robotics Perception and Intelligence, the Dept. of Electronic and Electrical Engineering, Southern University of Science and Technology, Shenzhen, China, and the Dept. of Electronic Engineering, The Chinese University of Hong Kong, Hong Kong, China.}
}

\begin{document}

\maketitle
\thispagestyle{empty}
\pagestyle{empty}

%%%%%%%%%%%%%%%%%%%%%%%%%%%%%%%%%%%%%%%%%%%%%%%%%%%%%%%%%%%%%%%%%%%%%%%%%%%%%%%%
\begin{abstract}
Robust and accurate 2D/3D registration, which aligns preoperative models with intraoperative images of the same anatomy, is crucial for successful interventional navigation. To mitigate the challenge of a limited field of view in single-image intraoperative scenarios, multi-view 2D/3D registration is required by leveraging multiple intraoperative images. In this paper, we propose a novel multi-view 2D/3D rigid registration approach comprising two stages. In the first stage, a combined loss function is designed, incorporating both the differences between predicted and ground-truth poses and the dissimilarities (e.g., normalized cross-correlation) between simulated and observed intraoperative images. More importantly, additional cross-view training loss terms are introduced for both pose and image losses to explicitly enforce cross-view constraints. In the second stage, test-time optimization is performed to refine the estimated poses from the coarse stage. Our method exploits the mutual constraints of multi-view projection poses to enhance the robustness of the registration process. The proposed framework achieves a mean target registration error (mTRE) of $0.79 \pm 2.17$ mm on six specimens from the DeepFluoro dataset, demonstrating superior performance compared to state-of-the-art registration algorithms.

% Give one example result here
\end{abstract}
% \def\abstractname
% \textbf{\textit{Index terms}}
% \begin{abstract}
% 2D/3D Registration, Multi-view Registration, Surgical Navigation, Digitally Reconstructed Radiographs (DRR).
% \end{abstract}
%%%%%%%%%%%%%%%%%%%%%%%%%%%%%%%%%%%%%%%%%%%%%%%%%%%%%%%%%%%%%%%%%%%%%%%%%%%%%%%%
\section{Introduction}
\input{sections/introduction}
\label{Introduction}

\section{Related Work}

In this section, we review X-ray image synthesis techniques and 2D/3D registration approaches, in the field of computer-assisted interventions.
\input{sections/relatedwork}
\label{relatedwork}

\section{Method}
\input{sections/method}
\label{Method}

\section{Experiments}
\input{sections/results}

\label{Results}

\section{Conclusions}
\input{sections/conclusion}
\label{Conclusion}

% \addtolength{\textheight}{-12cm}   
% \section*{ACKNOWLEDGMENT}
% This work was supported by the National Natural Science Foundation of China under Grant 62303275, the National Natural Science Fund for Excellent Young Scientists Fund Program (Overseas) under Grant 22IAA01849, Jinan Science and Technology Bureau under Grant 202333011.

\bibliographystyle{IEEEtran}
\bibliography{root}

\end{document}

%% file: sections/introduction.tex
Various medical imaging modalities, such as magnetic resonance imaging (MRI) \cite{He2024}, ultrasound \cite{Chan2022}, computed
 tomography (CT) \cite{Albano2024}, X-ray, and endoscopy \cite{Kwok2024}, are widely used to visualize internal anatomy (e.g., blood vessels \cite{Rossitti2009}) and guide surgical navigation (e.g., periacetabular osteotomy \cite{Unberath2021}). Interventional radiology, now being a standard in image-guided surgery, typically involves X-ray-based navigation (e.g., spinal needle injection) to estimate the 3D pose of surgical tools relative to anatomical structures (usually preoperative CT) from 2D X-ray images \cite{Wang2017}. In orthopedic navigation, the goal of 2D/3D registration is to determine the optimal transformation aligning 3D representations (e.g., CT) with 2D observations (e.g., X-ray). Robust and accurate registration is crucial for advanced augmented-reality (AR) assisted interventions \cite{Gopalakrishnan2024, Cho2023}.

Despite extensive studies on 2D/3D medical image registration, the development of an effective end-to-end automatic method remains an active research area \cite{Zhang2023, Unberath2021}. Traditional X-ray to CT registration methods often rely on iterative optimization. Landmark-based point-to-point registration \cite{Groher2007} requires manual labeling, limiting its direct suitability for minimally invasive procedures \cite{Jiao2023}. The DRR-based (Digitally Reconstructed Radiographs) method aims to maximize similarity metrics, such as normalized cross-correlation (NCC) or structural similarity (SSIM), between synthetic and real X-rays \cite{Esfandiari2019, Gopalakrishnan2022}, in order to optimize pose parameters for matching synthetic and real X-rays. However, they suffer from limited capture range and are sensitive to initial poses \cite{Unberath2021}.

\begin{figure}[t] 
\centering 
\includegraphics[page=1,width=0.47\textwidth]{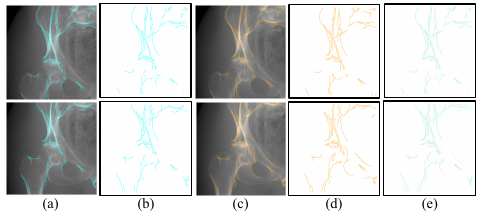} 
\vspace{-0.4cm}
\caption{\textbf{DRRs with contour.} Two rows display the DRR images and contour information in different poses, showing the projections in estimated and true poses after fine registration. In Column (a), X-ray images are rendered under two true poses, with contours in \textcolor{cyan}{Cyan}. Column (b) retains only the contours from (a). Column (c) and (d) show the X-ray images and contours under estimated poses, with contours in \textcolor{orange}{Orange}. Column (e) overlays (b) and (d), where overlapping regions in \textcolor{light_aquamarine}{Light Aquamarine} indicate higher accuracy.} 
\label{edge} 
\vspace{-0.3cm}
\end{figure}

Recent advances in deep learning aim to replace traditional iterative optimization process with trainable models owning a large number of learnable parameters. In landmark-based registration, feature extractors establish the mapping relationship between 3D anatomical landmarks and 2D image landmarks and a Perspective-n-Point (PnP) solver \cite{Lepetit2009} estimates the 2D image's pose relative to the 3D structure \cite{Shrestha2023, Grupp2020}. Although this approach shows promising results, its reliance on manually labeled data limits its real-world surgical applicability \cite{Bier2018, Grupp2019}. Consequently, recent studies focus on unsupervised or self-supervised registration methods, estimating the X-ray's pose relative to the anatomy through a single inference process \cite{Zhang2023, Gopalakrishnan2024}.

\begin{figure*}[htbp]
\centering
\includegraphics[page=1,width=\textwidth]{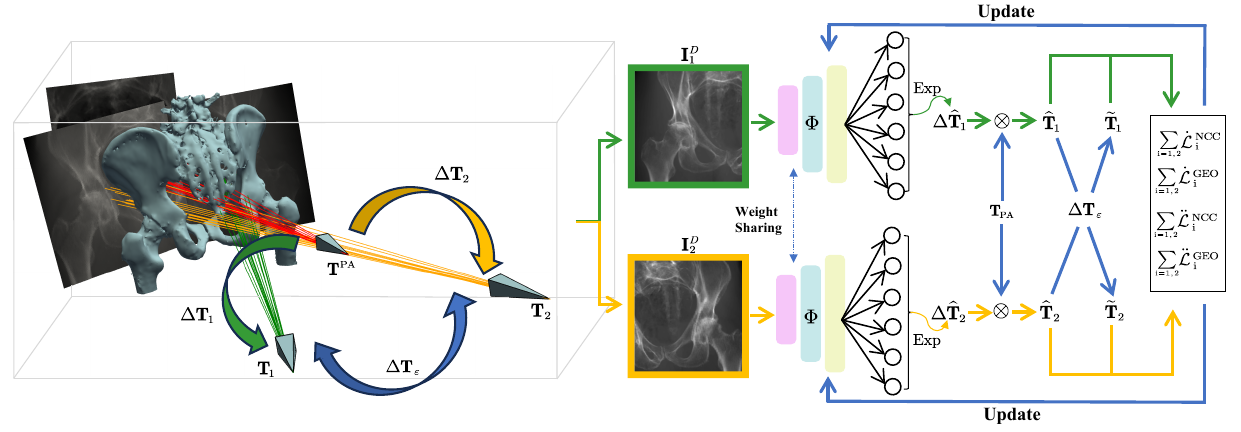}
\vspace{-0.8cm}
\caption{\textbf{The proposed framework during the training process.} The box on the left represents a DRR renderer, where the pyramids symbolizes the sampled camera poses. The right side depicts the training process of the pose estimation network, with the \textcolor{green}{Sea Green} loop representing the training process for Viewpoint 1, the \textcolor{yellow}{Amber} loop for Viewpoint 2, and the \textcolor{blue}{Steel Blue} loop for cross-constraints and parameter updates.}
\label{new_AP}
\vspace{-0.5cm}
\end{figure*}
%While single-image inference achieves high 3D accuracy, 
The projection process in the physical formulation of X-ray images actually leads to a loss of crucial 3D information \cite{Mi2023} and the limited field of view \cite{Schaffert2019}, makes single-view registration an ill-posed problem, which leads to significant variability of registration accuracy (e.g., high standard deviation error). In this paper, based on the state-of-the-art DiffPose framework \cite{Gopalakrishnan2024}, we propose a novel robust and accurate multi-view 2D/3D rigid registration approach, where we introduce differentiable perturbations across multiple intraoperative views and establish cross-view constraints in both camera pose and image spaces. Our approach compensates for information loss and enhances pose mapping from perspective images to anatomical structures.

The contributions of this paper are summarized as follows: 
\begin{itemize} 
\item We first formally define the multi-view 2D/3D registration problem, which is then solved by utilizing a Lie algebra Log-Exp mapping and applying differentiable perturbations that generate multiple intraoperative views.
\item We carefully leverage cross-view constraints (which can be formulated between every two intraoperative views) in both poses differences and image similarities, to formulate the additional training loss terms.
\item We test our framework on the DeepFluoro dataset under two challenging scenarios, demonstrating superior performance across all six specimens.
% dynamic 3D scenes with a constant view and static 3D scenes with varying spatial positions
\end{itemize}

%% file: sections/relatedwork.tex
\subsection{DRR X-ray Image Synthesis}  
Digitally Reconstructed Radiographs (DRR) generate simulated X-ray images from volumetric data, modeling ray attenuation to visualize anatomical structures \cite{Schmitt2024}. Recent advances in deep learning and GPU technology have significantly enhanced DRR performance. DeepDRR utilizes convolutional neural networks to decompose CT volumes and estimate Rayleigh scattering for realistic X-ray generation \cite{Unberath2018}. Gopalakrishnan et al. reformulates Siddon's method with vector-tensor operations for fast, differentiable DRR generation \cite{Gopalakrishnan2022}.
CT2X-GAN reduces style differences between DRR and real X-rays using a content-disentangled encoder \cite{Tan2024}.

\subsection{2D/3D Registration Approaches}\textbf{Optimization-Based 2D/3D Registration} The advancements in DRR technology have led to significant research focused on maximizing the similarity between synthesized and real X-ray images through pose optimization for the 2D/3D registration problem \cite{Berger2016, Liu2022}. Differentiable renderers (DiffDRR) overcome the non-differentiable nature of traditional renderers \cite{Grupp2019}, driving rapid estimation of the actual pose of intraoperative X-rays relative to preoperative CT structures using gradient-based optimization \cite{Gopalakrishnan2022}, demonstrating strong convergence properties.
% \subsection{Optimization-Based 2D/3D Registration}
% The advancements in DRR technology have led to significant research focused on maximizing the similarity between synthesized and real X-ray images through pose optimization for 2D/3D registration \cite{Berger2016, Liu2022}. Differentiable renderers (DiffDRR) overcome the non-differentiable nature of traditional renderers \cite{Grupp2019}, driving rapid estimation of the actual pose of intraoperative X-rays relative to preoperative CT structures using gradient-based optimization \cite{Gopalakrishnan2022}, demonstrating strong convergence properties. Zhang et al. proposed a method for computing the zero-normalized cross-correlation (ZNCC) between synthesized and real X-rays, aligning camera poses by maximizing the edge and structural information of the two images \cite{Zhang2023}.
\begin{figure*}[htbp]
\centering
\includegraphics[page=1,width=\textwidth]{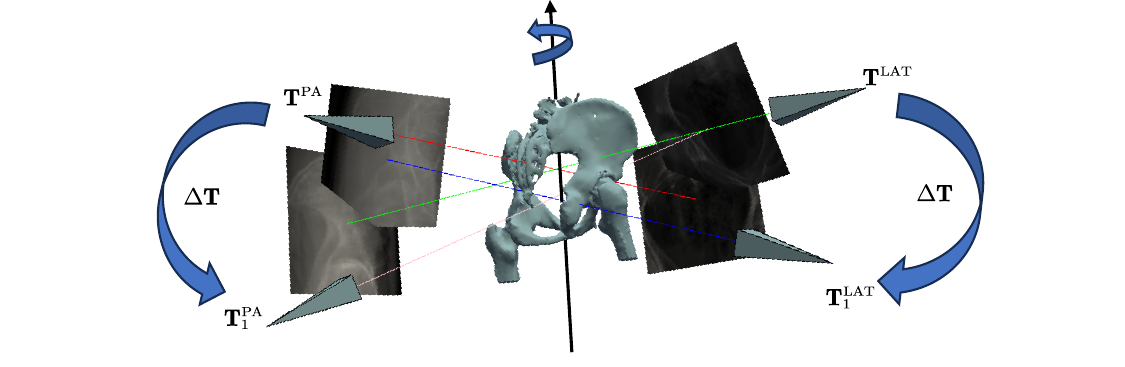}
\vspace{-0.8cm}
\caption{\textbf{Renderings from different viewpoints for a specific 3D scene.} The pyramids represent the cameras, with the \textcolor{LATblue}{Steel Blue} arrow indicating a twist around an axis passing through the CT center in the 3D space. The X-ray images in the figure are projections from PA and LAT views and after undergoing a simultaneous twist, respectively.}
\label{AP_LAT}
\vspace{-0.5cm}
\end{figure*}

\textbf{Learning-Based 2D/3D Registration} Some methods use 3D and 2D anatomical landmarks to estimate camera pose via a PnP solver. Grupp et al. employed a U-Net-based segmentation network to extract predefined landmark positions from X-ray segmentation masks \cite{Nguyen2022, Grupp2020}. However, this approach struggles when few projected landmarks are available, increasing annotation costs. In contrast, Gopalakrishnan et al. used DRR-synthesized X-rays to train a self-supervised patient-specific registration framework, eliminating the need for manual annotations \cite{Gopalakrishnan2024}. Jaganathan et al. improved registration robustness by combining domain adaptation with unsupervised features and style transfer to address X-ray style variations \cite{Jaganathan2023}.

\textbf{Multi-View 2D/3D Registration} To address the challenges of 2D/3D registration in cases of limited field of view or small structures, researchers have begun exploring multi-view registration methods. Liao et al. proposed $POINT^2$, which establishes point-to-point correspondences by tracking a set of interest points from CT projected onto DRRs in multiple views, directly aligning 3D data with the patient, and driving shape alignment between the patient and 3D data by aligning the 3D matching points \cite{Liao2019}. Schaffert et al. extended the point-to-plane correspondence (PPC) model, generalizing the optimization process to a view-independent coordinate system to perform 2D/3D registration across multiple viewing angles \cite{Schaffert2017}.

%% file: sections/method.tex
\subsection{Problem Formulation}
Let $ \mathbf{V} : \mathbb{R}^{3} \rightarrow \mathbb{R}$ denote the preoperative CT volume of one patient, $\mathbf{s}\in \mathbb{R}^{3}$ represents the position of the intraoperative X-ray source, and $\mathbf{g}\in \mathbb{R}^{3}$ represents the target pixel position on the detector plane. By ignoring complex reflection and scattering modes present in real X-ray systems and considering only linear attenuation, the energy attenuation $\bar{E}(\mathbf{g})\in\mathbb{R}$ at $\mathbf{g}$ is defined as follows \cite{Gopalakrishnan2022}:
\begin{equation}
\bar{E}(\mathbf{g}) = \Vert{\mathbf{g}-\mathbf{s}}\Vert_2\sum_{m=1}^{M-1}(\alpha_{m+1}-\alpha_{m})\mathbf{V}[\mathbf{s}+\frac{\alpha_{m+1}+\alpha_{m}}{2}(\mathbf{g}-\mathbf{s})],
\label{energydecay}
\end{equation}
where $\alpha_m\in\mathbb{R}$ parameterizes the locations where the ray intersects one of the orthogonal planes comprising the CT volume, and $M\in \mathbb{N}^{+}$ is the number of intersected voxels. The energy of the rendered X-ray $\mathbf{I}^\text{D} : \mathbb{R}^{2} \rightarrow \mathbb{R}$ can then be described by the Beer-Lambert law \cite{swinehart1962}, where D represents the X-ray image generated by the DRR:
\begin{equation}
E(\mathbf{g}) = E_0\exp^{-\bar{E}(\mathbf{g})},
\label{Beer-Lambert}
\end{equation}
which is defined as a rendering operator $\mathscr{R}$. Given an intrinsic matrix $\mathbf{K}\in \mathbb{R}^{3 \times 4}$ for the camera, and an extrinsic matrix $\mathbf{T_\text{i}}\in \text{SE(3)}$ with $\ \text{i} \in \mathbb{N}^+$, where SE(3) is Special Euclidean group in three dimensions and i represents the i-th camera pose, the rendering operator takes a CT volume image and an extrinsic matrix as inputs and outputs the corresponding rendered X-ray image:
\begin{equation}
\mathbf{I^\text{D}_\text{i}} = \mathscr{R}(\mathbf{V}, \mathbf{T_\text{i}}),
\label{render}
\end{equation}
where $\mathbf{T_\text{i}}\in \text{SE}(3)$ belongs to the Lie group, and is composed of a rotation matrix $\mathbf{R_\text{i}}\in \text{SO(3)}$ and a translation vector $\mathbf{t_\text{i}}^\top \in \mathbb{R}^{3}$, 
\begin{equation}
\mathbf{T_\text{i}} = \begin{bmatrix} \mathbf{R_\text{i}} & \mathbf{t_\text{i}} \\ \mathbf{0}^{\mathsf{T}} & 1 \end{bmatrix}
% \in \text{SE(3)}.
\label{T_i}
\end{equation}
The 2D/3D registration problem involves finding the best set of projective transformations (i.e., rigid transformation) and non-rigid transformation (e.g., affine transformation) to align the 3D anatomical structures of an object with its 2D observations, given a real X-ray image $\mathbf{I}^\text{T} : \mathbb{R}^{2} \rightarrow \mathbb{R}$, where T represents the real X-ray image, or a simulated X-ray image $\mathbf{I}^\text{D}$ \cite{Unberath2021}. Given a set of 3D datasets $\mathbf{V}_\text{j} : \mathbb{R}^{3} \rightarrow \mathbb{R}$ and 2D projection data $\mathbf{I}^\text{j}_\text{i} : \mathbb{R}^{2} \rightarrow \mathbb{R}$, where $\text{j}\in\mathbb{N}^{+}$ represents the j-th patient, the optimization problem for general rigid registration can be formulated as:
\begin{equation}
\{\widehat{\mathbf{K}}^\text{j},\widehat{\mathbf{T}}^\text{j}_\text{i}\} = \underset{{\mathbf{K}^\text{j},\mathbf{T}^\text{j}_\text{i}}}{\text{arg min}} \sum_\text{i, j}C(\mathbf{K}^\text{j}\mathbf{T}^\text{j}_\text{i}(\mathbf{V}^\text{j}),\mathbf{I}^\text{j}_\text{i})+\Omega
\label{problem}
\end{equation}
where $\widehat{\mathbf{K}}^\text{j}\in \mathbb{R}^{3 \times 4}$ and $\ \mathbf{K}^\text{j} \in \mathbb{R}^{3 \times 4}$ are the estimated and true intrinsic camera matrices, and $\widehat{\mathbf{T}}^\text{j}_\text{i}\in\text{SE(3)}$ and $\mathbf{T}^\text{j}_\text{i}\in\text{SE(3)}$ are the estimated and true extrinsic camera matrices for the j-th patient in the i-th view, $C$ represents the cost function (e.g., similarity metric for images), and $\Omega$ denotes the regularization term. It is noteworthy that when i=1, the problem simplifies to single-view registration. Conversely, when i$\geq$1, the problem generalizes to multi-view registration.
% \begin{figure*}[htbp]
% \centering
% \includegraphics[page=1,width=\textwidth]{pdf/AP_LAT.pdf}
% \caption{2D/3D registration from different angles for a specific 3D scene. The cones represent the cameras, with the \textcolor{LATblue}{Steel Blue} arrow indicating a twist around an axis passing through the CT center in the 3D space.}
% \label{AP_LAT}
% \end{figure*}

\subsection{Multi-View Registration for Temporal Dynamics}\label{3B}
\subsubsection{Training the Pose Regressor Model}
  In this context, registration is performed using multiple X-ray frames generated near a fixed angle, such as the Postero-Anterio (PA) view. Without loss of generality, we consider two intraoperative X-rays but can be easily extended to clinical scenarios where multiple intraoperative views are captured. As shown in Figure \ref{new_AP}, for a given CT volume $\mathbf{V}$, we sample poses near the typical Postero-Anterio (PA) view. Since the extrinsic matrix $\mathbf{T_\text{i}}$ lies on the non-linear SE(3) manifold \cite{Jiang2024}, to perform dense sampling with gradients, we map the tangent space of SE(3) to a linear space using the associated Lie algebra $\mathfrak{se}(3)$. The logarithmic map 
  \begin{equation}
\text{Log}:\text{SE(3)}\rightarrow\mathbb{R}^6
  \end{equation}
  converts any transformation in SE(3) into a 6D vector. Conversely, the exponential map defined as 
  \begin{equation}
  \label{definition of Exp}
\text{Exp}:\mathbb{R}^6\rightarrow\text{SE(3)} 
  \end{equation}
  converts a corresponding 6D vector back into the SE(3) manifold. Therefore, we can sample the first viewpoint $\varepsilon_\text{1}\in\mathbb{R}^6$ from a Gaussian distribution $\mathcal{N}(^\text{d}\mu,^\text{d}\sigma^2)$ over $\mathbb{R}^6$, where $^\text{d}\mu\in\mathbb{R}$ and $^\text{d}\sigma\in\mathbb{R}$ are the mean and variance of the distribution for the $d$-th dimension. Using Exp$(\varepsilon_\text{1})$, we obtain a pose sample $\Delta\mathbf{T}_1$. To simulate multi-frame continuous perturbations around the typical view, further sampling is conducted based on $\varepsilon_1$. Therefore, we define the twist $\varepsilon\in\mathbb{R}^6$ from a Gaussian distribution $(0,2\cdot^d\sigma^2)$ over $\mathbb{R}^6$. The $\mathfrak{se}(3)$ vector of another view can be expressed as:
$\varepsilon_\text{2} = \varepsilon_\text{1} + \varepsilon ,\ \varepsilon_\text{2} \in\mathbb{R}^6$.
Similarly, Exp$(\varepsilon_2) = \Delta\mathbf{T}_2$. Thus, given the extrinsic matrix $\mathbf{T}^\text{PA}\in \text{SE(3)}$ for the typical Postero-Anterio view, the extrinsic matrices for two views are represented as: $\mathbf{T}_1 = \Delta\mathbf{T}_1\cdot\mathbf{T}^\text{PA}$ and $\mathbf{T}_2 = \Delta\mathbf{T}_2\cdot\mathbf{T}^\text{PA}$.

After obtaining the preoperative CT volume $\mathbf{V}$, we repeat the above process to sample camera poses and acquire the simulated X-ray dataset $\mathcal{I}^\text{D}=\{\mathscr{R}(\mathbf{V}, \mathbf{T_\text{i}})\}_\text{i=1}^\text{2}$ via the DRR process. We then train a pose estimation network $\Phi:\mathcal{I}\rightarrow\mathbb{R}^6$. For the two-frame images, we estimate two perturbations: $\Delta\widehat{\mathbf{T}}_\text{i} = \Phi(\mathbf{I^\text{D}_\text{i}}),\text{i=\{1, 2}\}$, and subsequently construct the extrinsic matrices for the corresponding two-frame images as $\widehat{\mathbf{T}}_1 = \Delta\widehat{\mathbf{T}}_1\cdot\widehat{\mathbf{T}}^\text{PA}$. Using the rendering operator $\mathscr{R}$, we render the X-ray images under the estimated poses: $\mathbf{\widehat{I}^\text{D}_\text{i}} = \{\mathscr{R}(\mathbf{V}, \widehat{\mathbf{T}}_\text{i})\}_\text{i=1}^\text{2}$. Finally, the loss function is constructed based on $\mathbf{T_\text{i}}$, $\widehat{\mathbf{T}}_\text{i}$, $\mathbf{I^\text{D}_\text{i}}$, $\mathbf{\widehat{I}^\text{D}_\text{i}}$, and $\varepsilon$, to optimize the weights of $\Phi$. 

\subsubsection{Cross Loss Function}
The loss function $\mathcal{L}\in\mathbb{R}$ consists of the local term $\dot{\mathcal{L}}\in\mathbb{R}$ and the cross term $\ddot{\mathcal{L}}\in\mathbb{R}$ as
\begin{equation}
\mathcal{L} = \beta_1\dot{\mathcal{L}} + \beta_2\ddot{\mathcal{L}}
\label{all_loss}
\end{equation}
where $\beta_1\in\mathbb{R}$, $\beta_2\in\mathbb{R}$ are hyperparameters controlling the two terms' contributions. Each term consists of two parts: one based on geodesic distance and the other on image similarity. \\
\indent \textbf{The Local Term} The local term $\dot{\mathcal{L}}$ is the sum of the geodesic distance $\mathcal{L}^{\text{GEO}}_{\text{i}}\in\mathbb{R}$ between the ground truth and predicted camera poses and the normalized cross-correlation (NCC) $\dot{\mathcal{L}}^{\text{NCC}}_{\text{i}}\in\mathbb{R}$ between the images projected under both poses:
\begin{equation}
\dot{\mathcal{L}} = \sum_{\text{i=1, 2}}(\gamma\dot{\mathcal{L}}^{\text{GEO}}_{\text{i}} + \dot{\mathcal{L}}^{\text{NCC}}_{\text{i}}),
\label{dot_loss}
\end{equation}
where $\gamma\in \mathbb{R}$ is a hyperparameter controlling their relative contributions.
More specifically, $\dot{\mathcal{L}}^{\text{GEO}}_{\text{i}}$ is the geodesic distance between ground-truth and predicted poses, defined as in \cite{Gopalakrishnan2024}:
\begin{equation}
\begin{aligned}
\dot{\mathcal{L}}^{\text{GEO}}_{\text{i}} = & %\sum_{\text{i}=1, 2} 
\sqrt{\frac{f^2_{\text{i}}}{4}\arccos^2\left(\frac{\text{Tr}(\mathbf{R_\text{i}}\mathbf{\widehat{R}_\text{i}})-1}{2}\right) + \Vert(\mathbf{t_\text{i}}-\mathbf{\widehat{t}_\text{i}})\Vert} \\
& + \Vert\text{Log}\big((\mathbf{T_\text{i}})^{-1}\mathbf{\widehat{T}_\text{i}}\big)\Vert,
\end{aligned}
\label{dot_geo}
\end{equation}
where $f_{\text{i}}\in\mathbb{R}^+$ represents the camera focal length, and $\text{Tr}(\cdot)$ denotes the trace of a matrix, $||\cdot||$ denotes L2 norm of a vector. For the X-ray image $\mathbf{I}_\text{i}$ (e.g., either real or simulated) under a given camera pose and the rendered X-ray image $\widehat{\mathbf{I}}_\text{i}$ using the predicted pose, $\dot{\mathcal{L}}^{\text{NCC}}_{\text{i}}$ represents the NCC loss between $\mathbf{I}_\text{i}$ and $\widehat{\mathbf{I}}_\text{i}$:
\begin{equation}
\dot{\mathcal{L}}^{\text{NCC}}_{\text{i}} =
%\sum_{\text{i}=1, 2}
\sum_{\mathbf{p}}\frac{\mathbf{I_\text{i}}(\mathbf{p})-\mathbf{\mu(I_\text{i})}}{\sigma(\mathbf{I}_\text{i})}\cdot\frac{\mathbf{\widehat{I}_\text{i}}(\mathbf{p})-\mathbf{\mu(\widehat{I}_\text{i})}}{\sigma(\mathbf{\widehat{I}}_\text{i})}, 
\label{dot_ncc}
\end{equation}
where $\mathbf{p}\in\mathbb{R}^2$ represents the pixel location, $\mu(\cdot)\in\mathbb{R}, \sigma(\cdot)\in\mathbb{R}$ are the mean and standard deviation of the image which taking $\mathbf{I}_\text{i}$ 
 as an example are defined as follows:
%as defined in Equation (\ref{mean_and_std}).
\begin{equation}
\mu(\mathbf{I}_\text{i}) = \frac{1}{P}\sum_{\mathbf{p}}\mathbf{I}_\text{i}(\mathbf{p}),\ \sigma(\mathbf{I}_\text{i}) = \sqrt{\frac{1}{P}\sum_{\mathbf{p}}[\mathbf{I}_\text{i}(\mathbf{p})-\mu(\mathbf{I}_\text{i})]^2}.
\label{mean_and_std}
\end{equation}
where $P\in\mathbb{N}^+$ is the total number of pixels in $\mathbf{I_\text{i}}$.

\indent \textbf{The Cross Term} Based on the perturbation $\varepsilon$, the pose estimation results from two frames can be used to perform cross-inference constraints twice. The cross loss term $\ddot{\mathcal{L}}$ involves cross constraints using $\varepsilon$ and is defined as:
\begin{equation}
\ddot{\mathcal{L}} = \sum_{\text{i=1, 2}}\gamma\ddot{\mathcal{L}}^{\text{GEO}}_{\text{i}} + \ddot{\mathcal{L}}^{\text{NCC}}_{\text{i}}.
\label{ddot_loss}
\end{equation}
where $\ddot{\mathcal{L}}^{\text{GEO}}_{\text{i}}\in\mathbb{R}$ represents the geodesic distance between the ground-truth pose and predicted poses using that of the other view (cf. Equation (\ref{the equation of getting the pose from the other view})), and the NCC cross-consistency loss $\ddot{\mathcal{L}}^{\text{NCC}}_{\text{i}}$ between one 2D intraoperative image and the renderd image using the pose of the other view (cf. Equation (\ref{ddot_ncc})). To achieve bi-directional constraints, we apply the exponential map defined in Equation (\ref{definition of Exp})
to $\varepsilon$:
\begin{equation}
\text{Exp} (\varepsilon)= \Delta\mathbf{\mathbf{T}}_\varepsilon = \begin{bmatrix} \mathbf{R_\varepsilon} & \mathbf{t_\varepsilon} \\ \mathbf{0} & 1 \end{bmatrix} \in \text{SE(3)}.
\label{Expvar}
\end{equation}
Given two estimated camera pose matrices, the twist $\varepsilon$ can be used to compute the deviation of one estimated pose from the other:
\begin{equation}
\label{the equation of getting the pose from the other view}
\widetilde{\mathbf{T}}_\text{1} = \text{Exp(Log(}\mathbf{T}_\text{2}\text{)}-\varepsilon\text{)},\ \widetilde{\mathbf{T}}_\text{2} = \text{Exp(Log(}\mathbf{T}_\text{1}\text{)}+\varepsilon\text{)}
\end{equation}
where $\widetilde{\mathbf{T}}_\text{i}$
contains $\widetilde{\mathbf{R}}_\text{i}$ and $\widetilde{\mathbf{t}}_\text{i}$ while 
$\mathbf{T}_\text{i}$ contains $\mathbf{R}_\text{i}$ and $\mathbf{t}_\text{i}$ for $i\in\{1,2\}$. 
The geodesic distance cross loss term $\ddot{\mathcal{L}}^{\text{GEO}}_{\text{i}}$, for $\text{i}=\{1, 2\}$, is defined as:
\begin{equation}
\begin{aligned}
% \sum_{\text{i=1, 2}}
\ddot{\mathcal{L}}^{\text{GEO}}_{\text{i}} = & \sqrt{\frac{f^2_{\text{1}}}{4}\arccos^2\left(\frac{\text{Tr}(\mathbf{R}_\text{i}\mathbf{\widetilde{R}_\text{i}})-1}{2}\right) + \Vert\mathbf{t_\text{i}}-\mathbf{\widetilde{t}_\text{i}}\Vert} \\
% + & \sqrt{\frac{f^2_{\text{2}}}{4}\arccos^2\left(\frac{\text{Tr}(\mathbf{R}_\text{2}\mathbf{\widetilde{R}_\text{2}})-1}{2}\right) + \Vert\mathbf{t_\text{2}}-\mathbf{\widetilde{t}_\text{2}}\Vert}
% \\
+ & \Vert\text{Log}(\mathbf{\mathbf{T}_\text{i} }\mathbf{\widetilde{T}_\text{i}})\Vert. %+ \Vert\text{Log}(\mathbf{\mathbf{T}_\text{2} }\mathbf{\widetilde{T}_\text{2}})\Vert.
\end{aligned}
\label{dot_geo}
\end{equation}

Using the twist parameter $\varepsilon$, we can render cross X-ray images, i.e., $\mathbf{\widehat{I}^\text{C}_\text{i}} = \mathscr{R}(\mathbf{V}, \mathbf{\widetilde{\mathbf{T}}_\text{i}})$, for $\text{i}=\{1, 2\}$. The NCC cross-consistency loss $\ddot{\mathcal{L}}^{\text{NCC}}_{\text{i}}$ penalizing the differences between $\mathbf{I}_\text{i}$ and $\mathbf{\widehat{I}_\text{i}^{\text{C}}}$ is then defined as 
\begin{equation}
\ddot{\mathcal{L}}^{\text{NCC}}_{\text{i}} = \sum_{\mathbf{p}}\frac{\mathbf{\widehat{I}_\text{i}}^{\text{C}}(\mathbf{p})-\mu(\mathbf{\widehat{I}_\text{i}^{\text{C}}})}{\sigma(\widehat{\mathbf{I}}_\text{i}^{\text{C}})}\cdot\frac{\mathbf{I_\text{i}}(\mathbf{p})-\mathbf{\mu(I_\text{i})}}{\sigma(\mathbf{I}_\text{i})}.
\label{ddot_ncc}
\end{equation}

\subsubsection{Intraoperative Fine Registration}\label{fine_regisration}
The aforementioned trained models often do not achieve perfect alignment, but the results of the pose regressor are usually near the optimal solution, which effectively addresses the limitation of the capture range. By integrating the real intraoperative X-rays and the preoperative CT, the registration results can be further refined using DRR techniques and image similarity metrics. Specifically, given two real intraoperative X-ray frames that are treated as the fixed images $\mathbf{I}_\text{i}$, based on the Lie algebra logarithmic transformation, the estimated camera pose $\widehat{\mathbf{T}}_\text{i}$ is considered differentiable. By rendering the X-ray images $\mathbf{\widehat{I}^\text{D}_\text{i}} = \mathscr{R}(\mathbf{V}, \mathbf{\widehat{T}_\text{i}})$ and calculating the similarity between $\mathbf{I}$ and $\mathbf{\widehat{I}^\text{D}_\text{i}}$ using Equation (\ref{dot_ncc}), $\widehat{\mathbf{T}}_\text{i}$ is updated based on the gradient to maximize the similarity between the fix image and the moving image.

\subsection{Multi-View Registration for Spatial Dynamics}\label{3C}
As illustrated in Figure \ref{AP_LAT}, for multiple views of a static 3D scene, a typical case would involve the Postero-Anterio view (PA) and the Lateral view (LAT). Under two fixed viewpoints, any motion of the 3D image can be viewed as a spatial transformation in the CT coordinate system. Therefore, for two fixed camera poses, the perturbations applied to the two sets of views are consistent. Given the prior knowledge of the relative transformation between camera viewpoints, we have:
\begin{equation}
\mathbf{T}^{\text{LAT}} =  \mathbf{T}^{\text{Trans}}\cdot\mathbf{T}^{\text{PA}},
\label{APtoLAT}
\end{equation}
Assuming the voxel size of the CT volume image in three dimensions is $\{x,y,z|x,y,z\in\mathbb{R}^+\},$ to ensure the rotation operation is always centered around the volume, $\mathbf{T}^{\text{Trans}}\in\text{SE(3)}$ is expressed as:
\begin{equation}
 \mathbf{T}^{\text{Trans}} = 
\begin{bmatrix}
\cos(\theta) & -\sin(\theta) & 0 & x + y\sin(\theta) \\
\sin(\theta) & \cos(\theta) & 0 & y - x\sin(\theta) \\
0 & 0 & 1 & 0 \\
0 & 0 & 0 & 1
\end{bmatrix},
\label{Trans}
\end{equation}
where $\theta\in[-\pi,\pi]$, which means the origin is first translated to the CT volume center, followed by a rotation operation, and then restored to an arbitrary origin. Without loss of generality, in our article, we primarily consider the LAT and PA views, with other atypical views being treated similarly. By setting $\theta=\frac{\pi}{2}$, we obtain:
\begin{equation}
 \mathbf{T}^{\text{Trans}} = 
 \begin{bmatrix}
0 & -1 & 0 & x + y \\
1 & 0 & 0 & y - x \\
0 & 0 & 1 & 0 \\
0 & 0 & 0 & 1
\end{bmatrix},
\label{Trans}
\end{equation}
In comparison to the method described in Section \ref{3B}, by letting $\Delta\mathbf{T}_{\varepsilon}= \mathbf{T}^{\text{Trans}}$, the registration for these two fixed-view scenarios can be elegantly handled using the approach outlined in Section \ref{3B}.

%% file: sections/results.tex
\subsection{Datasets}
We trained and evaluated the method from Section \ref{3B} using the publicly available DeepFluoro dataset \cite{Nguyen2022}, which includes hip CT scans from six cadavers (three male, three female), aged 57 to 94 years. 3D landmarks were manually digitized in the CT scans. On the left side, these landmarks include the left femoral head (L.FH), greater sciatic notch (L.GSN), inferior obturator foramen (L.IOF), major obturator foramen (L.MOF), superior pubic symphysis (L.SPS), inferior pubic symphysis (L.IPS), and Postero-Anterio iliac spine (L.ASIS). The corresponding landmarks on the right side are abbreviated as R.FH, R.GSN, R.IOF, R.MOF, R.SPS, R.IPS, and R.ASIS. A total of 366 real X-ray images were captured across the six CT scans, with recorded camera poses. During training, for each CT, 400,000 samples were generated using the DRR renderer $\mathscr{R}$, simulating 200,000 samples and 200,000 twist perturbations. Evaluation was conducted using 26 to 110 real X-rays per CT scan.

\subsection{Implementation Details}

We use ResNet18 and a two-layer MLP as the pose regressor network \cite{Gopalakrishnan2024} to estimate a 6D vector in $\mathfrak{se}(3)$. Experiments were conducted on an RTX 4090 Ti GPU, with the model trained using Adam at a learning rate of $10^{-3}$. DRR and real X-ray styles were aligned by iterating over maximum intensity values \cite{Gopalakrishnan2024}. The hyperparameters were set as $\beta_1=0.7$, $\beta_2=0.3$, and $\gamma=10^{-2}$. During fine registration, we used separate learning rates for the rotational ($7\times10^{-3}$) and translational ($7$) components. To ensure that images with lower NCC loss (indicating a significant deviation from the real pose-projected image) receive more focus, the cross-correlation information from Equation (\ref{dot_ncc}) is compared for the two views in each iteration. Lower NCC loss regions were assigned a weight of 0.8, and higher loss regions a weight of 0.2, with optimization over 500 iterations.
\vspace{-0.7cm}
\begin{figure}[h]
\centering
\includegraphics[page=1,width=0.45\textwidth]{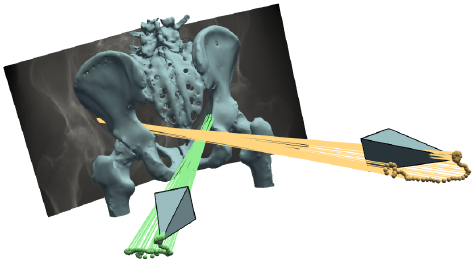}
\vspace{-0.3cm}
\caption{\textbf{Intraoperative fine registration.} The cones represent the camera poses in two frames of actual pose, while the positions occupied by the \textcolor{light_green}{Light Green} and \textcolor{Jasmine}{Jasmine} spheres denote the search space of the optimized camera poses for the two frames.}
\label{optimization}
\vspace{-0.3cm}
\end{figure}

\subsection{Evaluation Metrics}
To evaluate the method in Section \ref{3B}, we used the standard metrics of mean Target Registration Error (mTRE) and Sub-Millimeter Registration Success Rate (SMRSR) to assess the performance of our framework on real X-rays. mTRE is defined as the mean Euclidean distance between corresponding projected points of the true 3D landmarks under the ground truth and the estimated poses in the dataset, formulated as:
\begin{equation}
\text{mTRE} = \frac{1}{N}(\lambda\sum_{\text{i=1}}^{2}\sum_{n=1}^{N}\Vert{ \mathbf{K} (\mathbf{T}_\text{i}-\mathbf{\widehat{T}}_\text{i})
\begin{bmatrix}
\mathbf{L}_n \\
1
\end{bmatrix}}\Vert),
\label{mTRE}
\end{equation}
where $\mathbf{L}_n\in\mathbb{R}^{3}$ denotes the true coordinates of the $n$-th landmark in the CT, $N\in\mathbb{N}^+$ is the total number of 3D landmarks, and $\lambda\in\mathbb{N}^+$ is the pixel spacing factor, set to 0.194 in this context. SMRSR is defined as the percentage of X-ray images with an mTRE of less than 1 mm, where $w\in\mathbb{N}^+$, is the number of X-rays achieving this accuracy and $W\in\mathbb{N}^+$ is the total number of X-ray images used for registration:
\begin{equation}
\text{SMRSR} = \frac{w}{W}\times100\%.
\label{GFR}
\end{equation}
For the experiments described in Section \ref{3C}, due to the lack of a large number of X-rays captured from two fixed angles and their perturbed images, we conducted experiments entirely on DRR images from a single CT scan to demonstrate the generalizability of the framework.
\subsection{Results}
\vspace{-0.3cm}
\begin{table}[htbp]
\centering
\caption{Results on Six Specimens Before Fine registration.}
\scalebox{1.1}{
\begin{tabular}{l|cc|cc}
	\cline{1-5}
    & \multicolumn{2}{c|}{\textbf{DiffPose} \cite{Gopalakrishnan2024}} & \multicolumn{2}{c}{\textbf{Ours}} \\
     \cline{1-5}
    \rule{0pt}{10pt}& mTRE (mm) & SMRSR & mTRE (mm) & SMRSR \\
    \cline{1-5}
    \textbf{CT1} & 6.30 $\pm$ 3.81 & 0$\%$ & $\textbf{5.48} \pm \textbf{3.28}\uparrow$ & 0$\%$ \\
    \textbf{CT2} & 4.32 $\pm$ 3.77 & 0.90$\%$ & $\textbf{2.91} \pm \textbf{1.96}\uparrow$ & $\textbf{5.45\%}\uparrow$ \\
    \textbf{CT3} & 5.28 $\pm$ 1.34 & 0$\%$ & $\textbf{5.08} \pm \textbf{1.64}\uparrow$ & 0$\%$ \\
    \textbf{CT4} & 6.69 $\pm$ 2.53 & 5$\%$ & $\textbf{6.06} \pm \textbf{2.64}\uparrow$ & 0$\%$ \\
    \textbf{CT5} & 5.31 $\pm$ 2.51 & 0.90$\%$ & $\textbf{4.56} \pm \textbf{1.90}\uparrow$ & \textbf{0.91}$\%\uparrow$ \\
    \textbf{CT6} & 9.39 $\pm$ 9.90 & 20$\%$ & $\textbf{9.10} \pm \textbf{9.26}\uparrow$ & 0$\%$ \\
    \cline{1-5}
    \textbf{Total} & 5.81 $\pm$ 4.30 & 0.55$\%$ & $\textbf{4.89} \pm \textbf{3.69}\uparrow$ & $\textbf{1.94}\%\uparrow$ \\
	% \specialrule{1.5pt}{0pt}{0pt}
\end{tabular}}
\label{mTRE_table}
\vspace{-0.4cm}
\begin{tablenotes}
\item[*] \begin{flushleft}
\item[*]The mTRE values in the table represent the mean and standard deviation. The best result for each row is shown in bold.
\end{flushleft}
\end{tablenotes}
\end{table}

% To validate the improvement in multi-angle registration accuracy achieved by our core framework, we compare it with the state-of-the-art method, DiffPose \cite{Gopalakrishnan2024}, on the same dataset. Table \ref{mTRE_table} reports the registration results for six datasets, using a single inference of the trained pose regressor network on real X-rays. All X-rays in the dataset are near the AP view, thus any two X-rays of a patient can be considered as generated through a set of random perturbations. Based on this, we randomly pair X-rays of the same patient for testing. By introducing cross constraints, we reduced the mean projection error across all six datasets.
\vspace{-0.1cm}
To validate the improvement in multi-angle registration accuracy achieved by our core framework, we compare it with the state-of-the-art method, DiffPose \cite{Gopalakrishnan2024}, on the same dataset. Table \ref{mTRE_table} reports the registration results for six specimens, using a single inference of the trained pose regressor network on real X-rays. Any two X-rays of a patient can be considered as generated through a set of random perturbations. Based on this, we randomly pair X-rays of the same patient for testing. By introducing cross constraints, we reduced the mean projection error across all.
\vspace{-0.6cm}
\begin{figure}[h]
\centering
\includegraphics[page=1,width=0.49\textwidth]{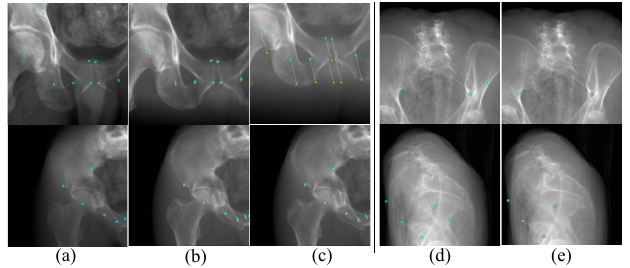}
\vspace{-0.7cm}
\caption{\textbf{Visualization of registration results.} The \textcolor{cyan}{Cyan} markers represent the 3D marker projections in the true pose, while the \textcolor{orange}{Orange} markers represent the 3D marker projections in the estimated pose. Ideally, the markers would overlap completely and appear \textcolor{light_aquamarine}{Light Aquamarine}. The mTRE values for column (b) are 0.35mm (Up) and 1.59mm (Down), for column (c) are 25.99mm (Up) and 0.35mm (Down),and for column (e) are 0.34mm(Up) and 0.66mm(Down) respectively.}
\label{compare}
\end{figure}
\vspace{-0.2cm}

Intraoperatively, we aim to refine registration through optimization by leveraging the similarity between images using DRR techniques. With the trained pose regressor network $\Phi$, we first perform a direct inference on two intraoperative X-ray frames, using the inference results as the initialization for the optimization process. This ensures that the initial pose estimates are as close as possible to the true pose, reducing the search space and avoiding local optima.

\vspace{-0.2cm}
\begin{table}[htbp]
\centering
\caption{Results on Six Specimens After Fine Registration.}
\scalebox{1.1}{
\begin{tabular}{l|cc|cc}
	\cline{1-5}
    & \multicolumn{2}{c|}{\textbf{DiffPose} \cite{Gopalakrishnan2024}}  & \multicolumn{2}{c}{\textbf{Ours}} \\
     \cline{1-5}
    \rule{0pt}{10pt}& mTRE (mm) & SMRSR & mTRE (mm) & SMRSR \\
    \cline{1-5}
    \textbf{CT1} & 0.92 $\pm$ 2.81 & \textbf{91$\%$} & $\textbf{0.73} \pm \textbf{1.49}\uparrow$ & 88$\%$ \\
    \textbf{CT2} & 1.10 $\pm$ 4.48 & 79$\%$ & $\textbf{0.66} \pm \textbf{1.13}\uparrow$ & $\textbf{87\%}\uparrow$ \\
    \textbf{CT3} & 0.50 $\pm$ 0.41 & 91$\%$ & $\textbf{0.50} \pm \textbf{0.31}\uparrow$ & 91$\%$ \\
    \textbf{CT4} & 1.08 $\pm$ 1.32 & \textbf{77$\%$} & $\textbf{1.04} \pm \textbf{0.92}\uparrow$ & 74$\%$ \\
    \textbf{CT5} & 0.43 $\pm$ 0.82 & 92$\%$ & $\textbf{0.32} \pm \textbf{0.27}\uparrow$ & \textbf{94}$\%\uparrow$ \\
    \textbf{CT6} & 2.70 $\pm$ 5.23 & \textbf{58}$\%$ & $\textbf{2.42} \pm \textbf{4.37}\uparrow$ & 52$\%$ \\
    \cline{1-5}
    \textbf{Total} & 1.00 $\pm$ 3.24 & 84$\%$ & $\textbf{0.79} \pm \textbf{2.17}\uparrow$ & $\textbf{86}\%\uparrow$ \\
\end{tabular}}
\label{mTRE_fine}
\vspace{-0.4cm}
\begin{tablenotes}
\item[*] \begin{flushleft}
\item[*] The mTRE values in the table represent the mean and standard deviation. The best result for each row is shown in bold.
    \end{flushleft}
\end{tablenotes}
\vspace{-0.6cm}
\end{table}

Figure \ref{optimization} illustrates the fine registration process between two views, where the trajectory of the sphere demonstrates the algorithm's convergence. Figure \ref{compare} visualizes the registration results of DiffPose (column c) and Ours (column b). For the two simultaneously input frames, our method demonstrates higher robustness. It also illustrates registration results under fixed PA and LAT views (column e), where training and testing were conducted solely on the DRR data from CT1, achieving sub-millimeter accuracy without the need for fine registration.  Using the inference results from Table \ref{mTRE_table}, we optimized registration on 366 random X-ray pairs, as described in Section \ref{fine_regisration}. Table \ref{mTRE_fine} shows that mean and variance of the target registration error across six specimens were reduced, highlighting our framework's robustness and scalability. In Figure \ref{edge}, DRR rendering and Canny edge detection \cite{McIlhagga2011} demonstrate alignment between true and estimated poses, where fully overlapping edges appear green. Notably, our model does not significantly improve sub-millimeter accuracy, potentially due to frame information fusion diverting attention from individual frames. Future work will explore segmentation-based methods using dice calculations to ensure optimization remains focused on specific frames during information fusion.

%% file: sections/conclusion.tex
% based on the latest advances in 2D/3D registration \cite{Gopalakrishnan2024}
In this paper, we propose a novel  multi-view registration framework with one perturbation and two cross-consistency constraints. To maximize the information correlation between different viewpoints in a fully unsupervised manner, we employ cross-view mutual constraints based on the pose estimation results from different perspectives, utilizing differentiable transformations on the SE(3) manifold. This framework effectively leverages the rotational consistency between multiple views, resulting in more accurate registration during surgery. In the current experiments, a large amount of DRR data is still needed for training, which may limit the applicability in actual surgical scenarios. However, fortunately, our core framework can be directly integrated into any unsupervised multi-angle pose regression method. Given that our framework is patient-specific, we aim to address this limitation in the future by leveraging generalized large models, which may enable us to fine-tune with only a small number of X-rays during surgery to achieve satisfactory registration results.